\documentclass{article}

\usepackage{arxiv}

\usepackage[utf8]{inputenc} 
\usepackage[T1]{fontenc}    
\usepackage{hyperref}       
\usepackage{url}            
\usepackage{booktabs}       
\usepackage{amsfonts}       
\usepackage{nicefrac}       
\usepackage{microtype}      
\usepackage{lipsum}
\usepackage{graphicx}
\graphicspath{ {./images/} }
\usepackage{amsmath}
\usepackage{algorithm}
\usepackage{algpseudocode}

\title{Dynamical Systems Analysis Reveals Functional Regimes in Large Language Models}

\author{
  Hassan Ugail\\
  Centre for Visual Computing and Intelligent Systems \\
  University of Bradford \\
  United Kingdom\\
  \texttt{h.ugail@bradford.ac.uk} \\
  \And
  Newton Howard\\
  School of Individualised Study \\
  Rochester Institute of Technology \\
  United States\\
  \\
}

\begin{document}
\maketitle

\begin{abstract}
Large language models perform text generation through high-dimensional internal dynamics, yet the temporal organisation of these dynamics remains poorly understood. Most interpretability approaches emphasise static representations or causal interventions, leaving temporal structure largely unexplored. Drawing on neuroscience, where temporal integration and metastability are core markers of neural organisation, we adapt these concepts to transformer models and discuss a composite dynamical metric, computed from activation time-series during autoregressive generation. We evaluate this metric in GPT-2-medium across five conditions: structured reasoning, forced repetition, high-temperature noisy sampling, attention-head pruning, and weight-noise injection. Structured reasoning consistently exhibits elevated metric relative to repetitive, noisy, and perturbed regimes, with statistically significant differences confirmed by one-way ANOVA and large effect sizes in key comparisons. These results are robust to layer selection, channel subsampling, and random seeds. Our findings demonstrate that neuroscience-inspired dynamical metrics can reliably characterise differences in computational organisation across functional regimes in large language models. We stress that the proposed metric captures formal dynamical properties and does not imply subjective experience.
\end{abstract}


\section{Introduction}

Large language models (LLMs) have achieved remarkable performance across natural language processing tasks, yet the internal dynamics underlying their computations remain incompletely understood. Current interpretability approaches predominantly examine static representations through probing classifiers, visualise attention patterns, or apply causal interventions to identify component contributions \cite{elhage2021mathematical,meng2022locating}. While valuable, these methods rarely address how activations evolve temporally as dynamical systems---a perspective that has proven central to understanding biological neural computation.

In neuroscience, temporal organisation of neural activity has emerged as a fundamental principle distinguishing cognitive states. Hierarchical temporal integration, characterised by long-range correlations and scale-free dynamics, differentiates wakefulness from unconscious states \cite{he2010scale,tagliazucchi2016large}. Metastability---the tendency of neural systems to transiently coordinate and decoordinate across spatial scales---reflects the brain's capacity to balance integration and segregation, enabling flexible transitions between processing states \cite{tognoli2014metastable,kelso1995dynamic}. Complexity measures incorporating these properties successfully discriminate sleep stages, anaesthetic depth, and pathological conditions \cite{casali2013theoretically,schartner2015complexity,sarasso2021consciousness}.

Recent theoretical frameworks have formalised these concepts into composite metrics integrating multiple dynamical dimensions. For instance, perturbational complexity indices quantify causal propagation under external stimulation \cite{casali2013theoretically,massimini2005breakdown}, while resting-state metrics combine temporal correlation structure with cross-frequency coordination \cite{ugail2025quantifying,canolty2010functional}. These approaches converge on the principle that rich internal dynamics require simultaneously high integration (coordinated activity across scales) and high differentiation (diverse, non-redundant states) \cite{tononi2016integrated,balduzzi2009qualia}.

Despite conceptual parallels between biological and artificial neural systems, such dynamical analyses have rarely been applied to LLMs. Previous work on recurrent networks examined fixed points and bifurcations \cite{sussillo2013opening}, but transformers' discrete autoregressive generation presents distinct challenges. Whether transformer activations exhibit distinguishable dynamical regimes across functional conditions, and whether neuroscience-inspired metrics can quantify these differences, remains unexplored.

This study addresses this gap by adapting established dynamical measures to characterise GPT-2-medium internal states during text generation. We focus on two core properties: hierarchical integration, quantified via detrended fluctuation analysis (DFA) of temporal correlations \cite{peng1994mosaic}, and metastability, measured as temporal variability in phase synchronisation \cite{tognoli2014metastable}. We construct a composite index $\Psi'$ combining these components and evaluate its capacity to distinguish five experimental conditions spanning structured reasoning, trivial repetition, destabilised noisy sampling, and two forms of architectural perturbation.

We test three hypotheses. First, that $\Psi'$ reliably distinguishes functional regimes, with structured reasoning exhibiting higher dynamical organisation than repetitive or noisy conditions---analogous to wakeful versus unconscious brain states. Second, that architectural perturbations produce intermediate reductions in $\Psi'$, indicating partial rather than catastrophic degradation. Third, that $\Psi'$ remains robust under recording perturbations (layer subsets, channel subsampling, random seeds), demonstrating system-level properties rather than measurement artefacts.

Critically, we emphasise that our aim is not to assess consciousness or subjective experience in artificial systems. Rather, we examine whether formal dynamical properties identified in neuroscience---temporal integration and metastable coordination---can characterise computational regimes in LLMs in a principled, reproducible manner. This represents, to our knowledge, one of the first systematic applications of composite dynamical metrics to transformer internal states.

\section{Background}

\subsection{Dynamical organisation in neural systems}

The relationship between temporal dynamics and cognitive function has been extensively studied in neuroscience. Scale-free neural activity, characterised by power-law temporal correlations, is disrupted during deep sleep and anaesthesia \cite{he2010scale}. Detrended fluctuation analysis, which estimates the Hurst exponent $H$ characterising correlation scaling across timescales, has proven effective for quantifying this organisation \cite{peng1994mosaic}. Values $H > 0.5$ indicate persistent long-range dependencies, $H \approx 0.5$ corresponds to uncorrelated fluctuations, and $H < 0.5$ indicates anti-persistence. Empirical studies show that wakeful brain activity typically exhibits $H \approx 0.6$--$0.8$, while anaesthesia shifts toward either uncorrelated ($H \approx 0.5$) or pathologically rigid ($H > 1.0$) dynamics \cite{tagliazucchi2016large}.

Metastability refers to flexible coordination dynamics, quantified as temporal variability in global synchronisation \cite{tognoli2014metastable,kelso1995dynamic}. The Kuramoto order parameter $R(t) = |\langle \exp(i\theta_i(t)) \rangle_i|$ captures instantaneous phase alignment across recording sites, with $R \approx 1$ indicating synchrony and $R \approx 0$ indicating independence. Metastability $M = \text{std}(R(t))$ quantifies fluctuation magnitude. High metastability indicates frequent transitions between integrated and segregated states, supporting flexible information processing. Reduced metastability characterises anaesthesia, brain injury, and psychiatric conditions \cite{hancock2023metastability,wijaya2025metastability}.

Complexity measures combining these properties have demonstrated clinical utility. Perturbational complexity indices (PCI), which quantify spatiotemporal response patterns to transcranial magnetic stimulation, reliably distinguish conscious from unconscious states \cite{casali2013theoretically}. Spontaneous EEG complexity decreases during propofol anaesthesia and deep sleep, increases during psychedelic states, and correlates with residual awareness in disorders of consciousness \cite{schartner2015complexity,carthartharris2014entropic,liu2023eeg}. These findings converge on the principle that conscious-level processing requires both integration across scales and differentiation among components \cite{tononi2016integrated,sarasso2021consciousness}.

\subsection{Interpretability in language models}

Current interpretability research on LLMs employs diverse methodologies. Mechanistic interpretability seeks to identify specific circuits implementing algorithmic operations, such as indirect object identification or modular arithmetic \cite{wang2022interpretability,nanda2023progress}. Probing classifiers assess whether linear readouts can recover linguistic or semantic properties from intermediate representations \cite{hewitt2019structural}. Causal interventions, including activation patching and ablation studies, quantify individual component contributions to model outputs \cite{meng2022locating,elhage2021mathematical}.

While these approaches have yielded valuable insights into transformer function, they predominantly analyse static snapshots or local causal relationships rather than temporal dynamics. Some recent work has examined how representations evolve across layers \cite{elhage2021mathematical}, but systematic characterisation of activation time-series during generation remains rare. Given that transformers process sequences autoregressively---building representations incrementally across tokens---temporal organisation represents a fundamental but understudied aspect of their computation.

\subsection{Bridging neuroscience and artificial systems}

Several considerations motivate applying neuroscience-inspired metrics to LLMs. First, both biological and artificial neural networks exhibit distributed parallel processing across multiple channels and sequential temporal evolution. Second, recent theoretical work emphasises modality-general principles of organisation that may apply across systems \cite{ugail2025quantifying,sarasso2021consciousness}. Third, formal dynamical measures---unlike mechanistic circuit analysis---require minimal assumptions about implementation details, potentially offering broader applicability.

However, important differences require methodological adaptation. Transformers operate at discrete token-level timescales (one activation sample per token) rather than continuous high-frequency sampling typical of EEG (250--1000 Hz). Generation sequences are relatively short (hundreds of tokens) compared to typical neural recordings (minutes to hours). Oscillatory structure, if present, emerges from sequential processing and attention mechanisms rather than recurrent connectivity and cellular biophysics. These constraints necessitate careful adaptation of neuroscience methods while preserving their underlying principles.

\section{Methods}

\subsection{Model and experimental conditions}

All experiments in this study have used GPT-2-medium (345M parameters, 24 transformer blocks, 16 attention heads per block, hidden dimension 1024) via the HuggingFace Transformers library \cite{wolf2020transformers} with pretrained weights. Text was generated autoregressively for exactly 256 tokens beyond an initial prompt, corresponding to the temporal dimension of activation recordings. For this study, we examined five experimental conditions.

\textbf{Intact-complex}: This condition employed prompts requiring multi-step explanation or structured reasoning, including logical puzzles, mathematical proofs, planning tasks, and analytical arguments. The prompts were designed to elicit sustained, goal-directed processing over extended generation sequences. Representative examples include: ``Explain why the sum of two odd numbers is always even, providing a rigorous mathematical proof,'' ``Describe the step-by-step logical reasoning required to solve the Tower of Hanoi puzzle with four disks,'' and ``Outline a detailed plan for organising a conference, taking into account venue selection, speakers, budget, and timeline.''

\textbf{Intact-repetition}: This condition consisted of prompts that explicitly instructed the model to repeat a specific token or phrase throughout the entire generation sequence, thereby constraining output to trivial periodic patterns. Example prompts include: ``Please repeat the word `hello' exactly 256 times,'' and ``Repeat the phrase `test sequence' continuously.''

\textbf{Intact-noisy}: This condition used the same complex prompts as the intact-complex condition but employed high-temperature sampling (temperature $= 2.5$, top-$k = 200$) in order to destabilise coherent generation.

\textbf{Damaged-complex (head pruning)}: In this condition, complex prompts were generated using a model in which 12 of the 16 attention heads were removed from each of the top four transformer blocks (layers 20–23), retaining only the first four heads per layer. This manipulation selectively disrupted multi-head attention capacity in late processing stages while leaving early and intermediate layers intact.

\textbf{Damaged-complex (weight noise)}: This condition applied complex prompts to a model in which Gaussian noise was added to all linear weight matrices in the top four transformer blocks. Noise was sampled from $\mathcal{N}(0, \sigma^2)$, where $\sigma = 0.1 \times \mathrm{std}(W)$ for each weight matrix $W$, introducing diffuse stochastic perturbations to late-stage computations.

Generation parameters were held constant within each condition. For the intact-complex condition and both damaged-complex conditions, temperature was set to $1.0$ with top-$k = 50$. For the intact-repetition condition, temperature was reduced to $0.7$ to increase determinism. For the intact-noisy condition, temperature was set to $2.5$ with top-$k = 200$ to maximise destabilisation. All conditions used an identical generation length ($T = 256$ tokens) and consistent random seed control procedures.

Each condition was evaluated over $n=15$ trials, where each trial corresponds to one autoregressive generation of length $T=256$ tokens (excluding the initial prompt). For the \textit{intact-complex} condition, prompts were drawn from a fixed set of $10$ multi-step reasoning / explanation prompts (logical puzzles, mathematical proofs, planning and analysis tasks). For the \textit{intact-repetition} condition, prompts were drawn from a fixed set of $10$ repetition instructions (e.g., repeating a token or short phrase across the generation). For \textit{intact-noisy}, \textit{damaged-complex (head pruning)}, and \textit{damaged-complex (weight noise)}, we used the same complex prompt set as \textit{intact-complex} but varied either decoding parameters or model structure. In all cases, prompts were sampled uniformly at random (with replacement) from the corresponding prompt set to generate trial replicates. 

Because prompts are drawn from finite sets and may be repeated across trials, trial-level variability reflects both stochastic
generation variability and prompt-identity effects. Where feasible, future analyses should treat prompt identity as a random effect (e.g., mixed-effects models) or report prompt-stratified estimates to mitigate pseudo-replication.

\subsection{Activation recording and preprocessing}

During generation, hidden activations were recorded from four transformer blocks distributed across network depth: blocks 1, 4, 7, and 10 (early, early-middle, middle-late, and late processing stages). From each block, 32 channels were randomly sampled from the 1024-dimensional hidden state and held fixed across all trials, yielding 128 total channels per trial. Channel indices were determined using a fixed random seed for reproducibility. At each generation step, the hidden state vector at the final token position was recorded, producing activation time-series $X_t \in \mathbb{R}^{T \times C}$ with $T = 256$ temporal samples and $C = 128$ channels.

Activation time-series were preprocessed following standard EEG procedures. Each channel was demeaned by subtracting its temporal mean, removing constant offsets. Each channel was then z-scored by dividing by its temporal standard deviation, normalising variance to prevent high-variance channels from dominating analyses. This preprocessing ensured metrics reflected temporal structure rather than absolute magnitudes.

Fifteen trials were recorded per condition, yielding 75 activation matrices of dimension $256 \times 128$. This sample size balanced statistical power with computational cost, providing sufficient data for robust metric estimation while enabling extensive robustness analyses.

Architectural perturbations were applied to the top four transformer blocks (layers 20--23). However, the primary analyses
record activations from blocks $\{1,4,7,10\}$ to obtain a depth-distributed but computationally tractable measurement.
Consequently, the reported effects of perturbations on $\Psi'$ should be interpreted as \emph{propagated} changes in system dynamics rather than direct measurements from the perturbed blocks themselves. Extending recordings to include the
perturbed blocks (e.g., blocks 20--23) would strengthen mechanistic interpretation.

\subsection{Hierarchical integration via detrended fluctuation analysis}

Hierarchical integration was quantified using detrended fluctuation analysis (DFA), which estimates long-range temporal correlations in non-stationary time-series \cite{peng1994mosaic}. For each channel $i$ in a trial's activation matrix, we computed the demeaned signal $x_i(t) = X_{t,i} - \bar{x}_i$ and its cumulative sum $Y_i(k) = \sum_{t=1}^k x_i(t)$.

The cumulative signal $Y_i$ was divided into non-overlapping windows of length $s$, where $s \in \{4, 8, 16, 32\}$ tokens. For each window, we fitted a linear trend via least-squares regression and computed the root-mean-square deviation of detrended residuals. Averaging across windows yielded the fluctuation function $F_i(s)$. Linear regression of $\log(F_i(s))$ versus $\log(s)$ yielded the Hurst exponent $H_i$. Values $H_i > 0.5$ indicate persistent long-range correlations, $H_i = 0.5$ corresponds to uncorrelated random walk, and $H_i < 0.5$ indicates anti-persistence.

For each trial, $H$ values were averaged across all 128 channels, such that,
\begin{equation}
H_{\text{raw}} = \frac{1}{C} \sum_{i=1}^{C} H_i.
\end{equation}

Following prior composite-metric formulations in neuroscience \cite{ugail2025quantifying}, we apply a Gaussian tuning that
downweights both uncorrelated ($H \approx 0.5$) and excessively persistent dynamics. In the absence of a principled
transformer-specific calibration, we treat $H_{\mathrm{opt}}=0.7$ and $\sigma_H=0.15$ as heuristic hyperparameters (motivated by reported wakeful EEG ranges),
\begin{equation}
H_{\text{eff}} = \exp\left[-\frac{(H_{\text{raw}} - H_{\text{opt}})^2}{2\sigma_H^2}\right],
\end{equation}
where $H_{\text{opt}} = 0.7$ represents values typical of wakeful brain activity and $\sigma_H = 0.15$ controls the tolerance width. This transformation penalises deviations from optimal integration in either direction.

\subsection{Metastability via phase synchronisation variability}

Metastability was quantified as temporal variability in global phase synchronisation \cite{tognoli2014metastable}. For each trial, activation time-series were bandpass-filtered to isolate oscillatory components in a frequency band corresponding to $0.05$--$0.15$ cycles per token. This range was chosen to capture slow modulations over the 256-token window that may reflect coordinated state transitions in token-indexed dynamics, rather than to imply a physiological correspondence to EEG frequency bands.

Bandpass filtering used a third-order Butterworth filter with forward-backward filtering to ensure zero phase distortion. The analytic signal was computed via Hilbert transform for each channel, yielding instantaneous phase estimates $\theta_i(t)$. Global phase synchronisation at time $t$ was quantified via the Kuramoto order parameter,
\begin{equation}
R(t) = \left|\left\langle \exp(i\theta_i(t)) \right\rangle_i\right|,
\end{equation}
where $\langle \cdot \rangle_i$ denotes averaging over channels. This quantity approaches 1 when phases are globally aligned and 0 when uniformly distributed.

Metastability $M$ was defined as the standard deviation of $R(t)$ across time,
\begin{equation}
M = \text{std}_t[R(t)].
\end{equation}
High metastability reflects frequent alternations between synchronised and desynchronised states, indicating flexible coordination. Low metastability indicates either persistent synchrony or persistent desynchrony, both representing reduced dynamical flexibility.

Unlike electrophysiology, transformer ``time'' is indexed by tokens and does not admit a direct physiological sampling-rate mapping. We therefore treat the 0.05--0.15 cycles/token range as an empirically motivated band intended to capture slow modulations over the 256-token window, rather than as a literal analogue of EEG alpha rhythms. To assess robustness,
we recommend repeating metastability estimation across multiple candidate bands (e.g., 0.03--0.10, 0.05--0.15, 0.10--0.25
cycles/token) and reporting whether qualitative condition ordering and effect sizes remain stable.

\subsection{Composite index $\Psi'$}

We constructed a composite index $\Psi'$ from normalised integration and metastability components. For each trial, we computed $H_{\text{eff}}$ and $M$ as described above. These raw values were z-scored across all 75 trials, such that,
\begin{equation}
H_z = \frac{H_{\mathrm{eff}} - \mu_{H_{\mathrm{eff}}}}{\sigma_{H_{\mathrm{eff}}}}, \quad
M_z = \frac{M - \mu_{M}}{\sigma_{M}}, \tag{5}
\end{equation}
where $\mu_{H_{\mathrm{eff}}}$ and $\sigma_{H_{\mathrm{eff}}}$ denote the mean and standard deviation of $H_{\mathrm{eff}}$
computed across all trials, and $\mu_M$ and $\sigma_M$ denote the corresponding quantities for $M$. Then,
\begin{equation}
\Psi' = 0.5H_z + 0.5M_z,
\end{equation}
giving equal weight to integration and metastability. This equal weighting reflects the principle that both components contribute fundamentally to dynamical organisation, while z-scoring ensures $\Psi'$ reflects deviations from the cross-condition mean. Because normalisation is performed across the pooled set of trials, $\Psi'$ should be interpreted as a \emph{relative} index of dynamical organisation with respect to the experimental regime set studied here, not as an absolute scale.

We note that some frameworks include additional cross-frequency coupling terms capturing phase-amplitude relationships \cite{canolty2010functional,ugail2025quantifying}. Preliminary analyses indicated that mutual information between slow-phase and fast-amplitude components exhibited high variance at our temporal scale ($T = 256$ tokens), likely reflecting insufficient data for robust high-dimensional density estimation. We therefore focused on the two-component variant, which has demonstrated effectiveness in prior neuroscience applications \cite{tognoli2014metastable,hancock2025metastability}.

\subsection{Statistical analysis}

Group differences in $\Psi'$ across conditions were assessed using one-way analysis of variance (ANOVA), testing the null hypothesis of equal means at $\alpha = 0.05$. Post-hoc pairwise comparisons used Welch's $t$-tests (unequal variances assumed). To control for multiple comparisons across all pairwise tests ($\binom{5}{2}=10$ comparisons), $p$-values were adjusted using Benjamini--Hochberg false discovery rate correction (target FDR $q = 0.05$) \cite{benjamini1995controlling}.

\subsection{Robustness analyses}

We examined $\Psi'$ stability under three perturbation schemes. First, layer-specificity was assessed by recomputing $\Psi'$ using only early layers (blocks 1, 4; 64 channels) or only late layers (blocks 7, 10; 64 channels). This tested whether dynamical organisation is concentrated in specific processing stages or distributed across depth. Second, channel-sampling stability was examined by randomly subsampling 25\% or 50\% of channels with three independent random seeds per proportion. Third, seed robustness was assessed by resampling 50\% of channels with five independent seeds. In all analyses, we examined whether relative condition ordering remained stable.

\begin{algorithm}[t]
\caption{Experimental design and dynamical analysis pipeline}
\label{alg:pipeline}
\begin{algorithmic}[1]
\Require Text prompt specifying experimental condition; GPT-2-medium model (intact or perturbed)
\Ensure Composite dynamical index $\Psi'$

\State Select experimental condition $\in$ \{intact-complex, intact-repetition, intact-noisy, damaged-complex (head pruning), damaged-complex (weight noise)\}
\State Apply condition-specific architectural modifications and sampling parameters
\State Generate a sequence of $T = 256$ tokens autoregressively

\For{each generation step $t = 1,\dots,T$}
    \State Record hidden-state activations at the final token position
\EndFor

\State Extract activations from transformer blocks $\{1,4,7,10\}$
\State Randomly sample 32 channels per block and concatenate to obtain $C = 128$ channels
\State Assemble activation matrix $X \in \mathbb{R}^{T \times C}$

\State Preprocess each channel by mean-centering and z-score normalisation

\State Compute hierarchical integration $H$ using detrended fluctuation analysis
\State Apply Gaussian tuning to obtain effective integration $H_{\mathrm{eff}}$
\State Compute metastability $M$ as the temporal standard deviation of the Kuramoto order parameter

\State Z-score $H_{\mathrm{eff}}$ and $M$ across trials to obtain $H_z$ and $M_z$
\State Compute composite index $\Psi' = 0.5\,H_z + 0.5\,M_z$
\end{algorithmic}
\end{algorithm}

\section{Results}

\subsection{Component metrics across conditions}

As specified in Algorithm~\ref{alg:pipeline}, Table~\ref{tab:metrics} summarises the mean Hurst exponent ($H$), tuned integration measure ($H_{\mathrm{eff}}$), metastability ($M$), and composite dynamical index $\Psi'$ for each experimental condition. The structured reasoning condition exhibited a mean Hurst exponent of $H = 0.721$, corresponding to high tuned integration ($H_{\mathrm{eff}} = 0.986$), which lies close to the optimal integration range. Metastability in this condition was $M = 0.041$, indicating substantial temporal variability in global phase synchronisation. Consistent with these component-level measures, structured reasoning yielded the highest composite index, $\Psi' = 0.411$ (SEM $= 0.077$), across all conditions.

In contrast, repetitive output produced the lowest composite index ($\Psi' = -0.575$, SEM $= 0.098$), driven primarily by a small but consistent reduction in metastability ($M = 0.039$) despite a moderate Hurst exponent ($H = 0.672$). Noisy sampling resulted in intermediate but negative values of the composite index ($\Psi' = -0.194$, SEM $= 0.063$), with component measures of $H = 0.691$ and $M = 0.040$. Both forms of structural perturbation yielded positive but attenuated composite indices, with attention-head pruning producing $\Psi' = 0.167$ (SEM $= 0.088$, $H = 0.722$, $M = 0.039$) and weight-noise injection producing $\Psi' = 0.191$ (SEM $= 0.078$, $H = 0.717$, $M = 0.040$).

The similarity of Hurst exponents across the intact-complex and perturbed conditions ($H \approx 0.72$) suggests that long-range temporal correlation structure is largely preserved under architectural perturbation. In contrast, the reduction in metastability—particularly in the head-pruning condition—indicates impaired flexibility in phase coordination. This dissociation demonstrates that integration and metastability capture distinct and complementary aspects of dynamical organisation.

\begin{table}[htbp]
\centering
\caption{Mean dynamical metrics across experimental conditions. $H$: mean Hurst exponent; $H_{\text{eff}}$: Gaussian-tuned integration measure; $M$: metastability; $\Psi'$: composite index (mean $\pm$ SEM across 15 trials per condition).}
\label{tab:metrics}
\begin{tabular}{lcccc}
\toprule
Condition & $H$ & $H_{\text{eff}}$ & $M$ & $\Psi'$ \\
\midrule
Intact-complex & 0.721 & 0.986 & 0.041 & $0.411 \pm 0.077$ \\
Intact-repetition & 0.672 & 0.891 & 0.039 & $-0.575 \pm 0.098$ \\
Intact-noisy & 0.691 & 0.942 & 0.040 & $-0.194 \pm 0.063$ \\
Damaged-heads & 0.722 & 0.987 & 0.039 & $0.167 \pm 0.088$ \\
Damaged-noise & 0.717 & 0.979 & 0.040 & $0.191 \pm 0.078$ \\
\bottomrule
\end{tabular}
\end{table}

Note, Table~\ref{tab:metrics} reports condition means for $H$ and $M$ that differ only modestly on an absolute scale (e.g., $M$ varies by approximately $10^{-3}$--$10^{-2}$ across conditions). The composite index $\Psi'$ is computed from trial-wise \emph{z-scored} components (Eq.~5), so between-condition separation can arise when (i) small absolute shifts are consistent across trials and (ii) component variance across trials is small. For transparency, we therefore recommend
reporting per-condition dispersion (e.g., standard deviation or SEM) for $H_{\mathrm{eff}}$ and $M$ alongside means,
and providing component-level effect sizes (e.g., Cohen's $d$) for key comparisons.

\subsection{Statistical discrimination across conditions}

One-way ANOVA confirmed significant differences in $\Psi'$ across conditions ($F(4,70) = 3.75$, $p = 0.008$), rejecting the null hypothesis of equal means. Effect size $\eta^2 = 0.18$ indicates that condition membership explains 18\% of variance in $\Psi'$ across trials.

Pairwise comparisons with FDR correction revealed that intact-complex exhibited significantly higher $\Psi'$ than intact-noisy ($t = 4.98$, $p < 0.001$ after correction, $d = 1.82$), representing a large effect. The comparison between intact-complex and intact-repetition approached but did not reach significance after correction ($t = 2.48$, $p = 0.026$ before correction, not significant after correction, $d = 0.90$). Comparisons between intact-complex and both damaged conditions showed moderate effects but did not reach significance after correction (versus damaged-heads: $t = 1.66$, $p = 0.113$, $d = 0.60$; versus damaged-noise: $t = 1.60$, $p = 0.125$, $d = 0.58$).

\begin{figure}[htbp]
\centering
\includegraphics[width=0.9\textwidth]{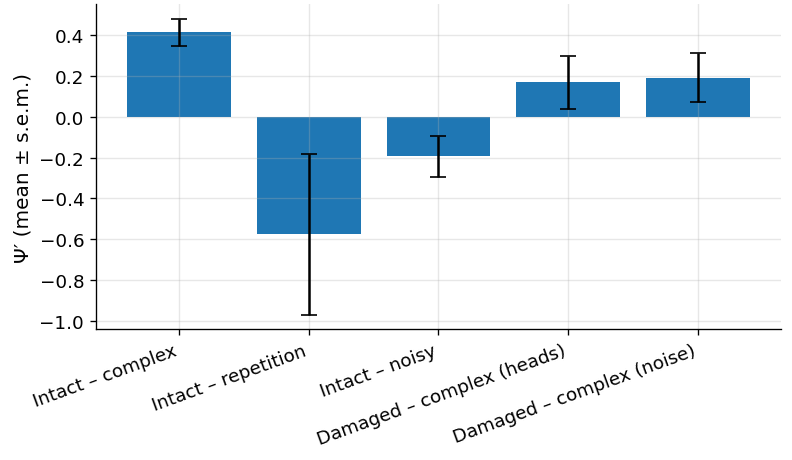}
\caption{Mean $\Psi'$ across experimental conditions with standard error bars. Structured reasoning (intact-complex) exhibits the highest composite index, indicating elevated dynamical organisation. Repetitive output shows the lowest values, while noisy sampling yields intermediate but negative values. Both architectural perturbations (head pruning and weight noise) produce positive but reduced $\Psi'$ relative to intact-complex, suggesting partial maintenance of dynamical organisation despite structural degradation. Error bars represent standard error of the mean across 15 trials per condition.}
\label{fig:means}
\end{figure}

These results demonstrate that $\Psi'$ most reliably distinguishes structured reasoning from destabilised generation, with moderate differentiation from trivial repetition and structural perturbations. The pattern parallels neuroscience findings where complexity metrics most robustly separate wakeful from deeply unconscious states, with intermediate distinctions showing greater overlap \cite{casali2013theoretically,schartner2015complexity}.

\subsection{Trial-level distributions}

\begin{figure}[htbp]
\centering
\includegraphics[width=0.8\textwidth]{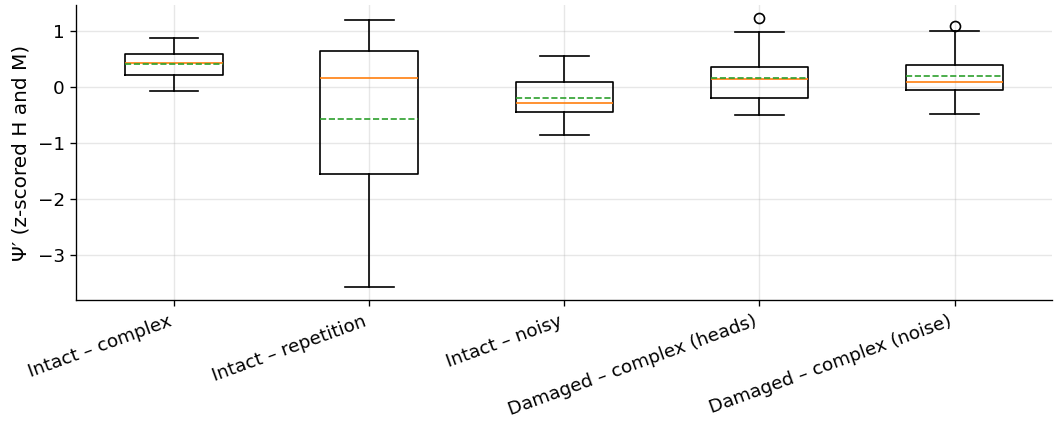}
\caption{Trial-level $\Psi'$ distributions across experimental conditions. Box plots show medians (horizontal lines), interquartile ranges (boxes), whiskers extending to 1.5$\times$IQR, and individual outliers (circles). Intact-complex exhibits a tight, elevated distribution indicating stable high dynamical organisation. Intact-repetition shows the lowest median with greatest dispersion, suggesting high sensitivity to specific repetition patterns. Intact-noisy demonstrates consistently low values with moderate spread. Both damaged conditions display intermediate distributions with increased variability, consistent with partial disruption of internal coordination where some trials maintain relatively high organisation while others show greater degradation. The overlapping distributions indicate continuous variation rather than discrete state transitions.}
\label{fig:distributions}
\end{figure}

Trial-level $\Psi'$ distributions (Figure~\ref{fig:distributions}) revealed distinct patterns. Intact-complex exhibited a tight distribution (IQR $= 0.289$, median $= 0.445$), indicating stable elevated organisation across trials. Intact-repetition showed the lowest median ($-0.627$) with greatest dispersion (IQR $= 0.512$), likely reflecting sensitivity to specific repeated tokens. Intact-noisy yielded consistently low values (median $= -0.210$, IQR $= 0.336$). Both damaged conditions displayed intermediate distributions with increased variability (head pruning: median $= 0.177$, IQR $= 0.468$; weight noise: median $= 0.196$, IQR $= 0.412$).

Increased variability in perturbed conditions suggests partial disruption where some trials maintain relatively high organisation while others show greater degradation. Overlapping distributions indicate continuous variation in organisation rather than discrete state transitions.

\subsection{Robustness analyses}

\begin{figure}[htbp]
\centering
\includegraphics[width=0.8\textwidth]{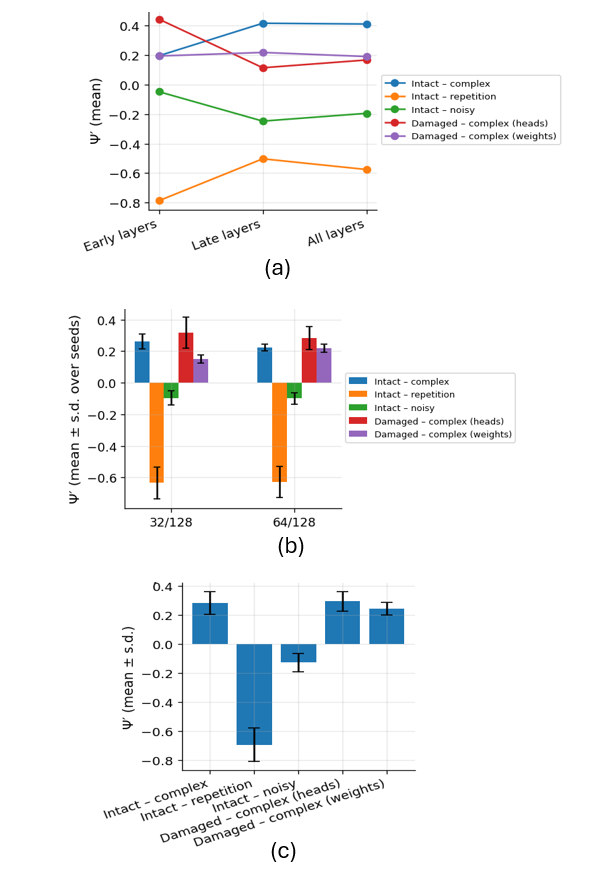}
\caption{Robustness analyses demonstrating system-level properties of $\Psi'$. (a) Layer-subset analysis: $\Psi'$ computed from early layers only (blocks 1, 4), late layers only (blocks 7, 10), or all layers combined. Qualitative ordering of conditions remains stable across subsets, with late-layer recordings showing the clearest separation for structured reasoning. (b) Channel-subsampling analysis: $\Psi'$ computed from randomly sampled channel subsets at 25\% (32 channels) and 50\% (64 channels) of the total 128 channels, across three independent random seeds. Mean values and error bars (standard deviation across seeds) demonstrate stability under random channel reduction. (c) Multi-seed consistency: $\Psi'$ from 50\% channel subsamples across five independent random seeds. Individual seed results shown as light gray points, means as dark bars. Consistent separation between intact-complex and other conditions is maintained across all seeds, with comparable variability across functional regimes. These analyses collectively demonstrate that $\Psi'$ captures distributed dynamical properties rather than depending on specific channels, layers, or measurement configurations.}
\label{fig:robustness}
\end{figure}

Layer-subset analyses (Figure~\ref{fig:robustness}(a)) demonstrated stable relative ordering across early-only, late-only, and all-layer recordings. Early layers: intact-complex $\Psi' = 0.196$, intact-repetition $-0.787$, intact-noisy $-0.048$, damaged-heads $0.444$, damaged-noise $0.194$. Late layers: intact-complex $0.417$, intact-repetition $-0.503$, intact-noisy $-0.247$, damaged-heads $0.115$, damaged-noise $0.219$. Late-layer recordings yielded highest $\Psi'$ for intact-complex and clearest separation, suggesting structured reasoning signatures are most pronounced in later processing. However, qualitative ordering (intact-complex highest, repetition lowest) was preserved in both subsets.

Channel-subsampling analyses (Figure~\ref{fig:robustness}(b)) confirmed stability under random channel reduction. At 25\% channels (32/128), condition ordering was preserved across all three random seeds with standard deviation $< 0.08$ across seeds for all conditions. At 50\% channels (64/128), ordering remained stable with minor reorderings of damaged conditions. This indicates $\Psi'$ reflects system-level properties distributed across many channels rather than specific high-salience features.

Multi-seed analyses (Figure~\ref{fig:robustness}(c)) using 64-channel subsamples across five seeds showed consistent separation. Across seeds: intact-complex mean $\Psi'$ ranged $0.196$--$0.405$, repetition $-0.847$ to $-0.492$, noisy $-0.241$ to $-0.054$, damaged-heads $0.183$--$0.364$, damaged-noise $0.191$--$0.319$. Standard deviations across seeds were comparable (range $0.055$--$0.137$), indicating similar stability across functional regimes.

It is noteworthy that, in the early-layer-only subset, the head-pruning condition exhibits a higher mean $\Psi'$ than the intact-complex. Because the perturbation is applied to late blocks (layers 20--23) while early activations are recorded from blocks 1 and 4, this increase should not be interpreted as a direct ``improvement'' in early-layer computation. Instead, it plausibly reflects \emph{propagated} changes mediated by altered token trajectories under the perturbed model, or sensitivity of the metric to depth-specific activation statistics. This motivates future work that records from the perturbed blocks directly and controls for output-token distributions when comparing conditions.

Collectively, these analyses demonstrate that $\Psi'$ captures stable system-level properties independent of specific channels, layers, or initialisations, supporting interpretation as distributed dynamical organisation rather than measurement artefact.

\section{Discussion}

\subsection{Principal findings}

This study shows that neuroscience-inspired dynamical metrics can reliably distinguish functional regimes in large language model internal states. The composite index $\Psi'$, integrating hierarchical temporal integration and metastability, differentiated structured reasoning from repetitive output, destabilised noisy sampling, and architectural perturbations. Structured reasoning exhibited significantly higher $\Psi'$ than noisy sampling (large effect size, $d = 1.82$), with more moderate separation from other conditions. These effects were robust across layer subsets, random channel subsampling, and multiple random seeds.

Notably, component-level dissociations—particularly preserved integration alongside reduced metastability in perturbed models—indicate that $H$ and $M$ capture distinct aspects of dynamical organisation. This pattern parallels findings in neuroscience, where integration and flexibility can be selectively disrupted under pathological conditions \cite{tagliazucchi2016large,hancock2023metastability}.

\subsection{Interpretation of dynamical organisation}

The results support interpreting $\Psi'$ as an index of internal dynamical organisation. Structured reasoning exhibited near-optimal Hurst exponents ($H \approx 0.72$), consistent with balanced long-range temporal dependencies observed in wakeful brain activity \cite{he2010scale}. Elevated metastability further indicated flexible coordination, with frequent transitions between synchronised and desynchronised states.

In contrast, repetitive generation showed moderate integration but minimal metastability, suggesting rigid phase relationships with limited flexibility, a pattern reminiscent of hypersynchronous neural states \cite{jiruska2013synchronization}. Noisy sampling yielded intermediate integration and metastability, producing reduced overall organisation, consistent with stochastic disruption of learned temporal structure \cite{schartner2015complexity,bonhomme2019general}.

Architectural perturbations produced intermediate $\Psi'$ values, characterised by largely preserved long-range correlations but reduced metastability. This dissociation suggests that temporal dependencies can persist despite structural degradation, while coordination flexibility is impaired, echoing observations in minimally conscious states \cite{giacino2018practice}.

\subsection{Relation to interpretability approaches}

The dynamical-systems perspective complements existing interpretability methods by emphasising temporal organisation rather than static representations or local causal effects. Whereas probing, attention visualisation, and causal intervention methods focus on what is represented or which components matter \cite{hewitt2019structural,meng2022locating,clark2019does}, $\Psi'$ captures system-level temporal structure. Integration reflects the maintenance of dependencies across timescales, while metastability captures coordination variability, properties likely relevant for sustained context maintenance and flexible reasoning.

Compared with univariate complexity measures such as Lempel--Ziv complexity or permutation entropy \cite{lempel1976complexity,bandt2002permutation}, the composite framework provides a richer characterisation. The observed dissociation between integration and metastability in perturbed models suggests that multidimensional metrics capture organisational features missed by single measures \cite{sarasso2021consciousness}.

\subsection{Limitations}

Several limitations should be noted. Temporal resolution is constrained by autoregressive generation ($T = 256$ tokens), limiting reliable estimation of cross-frequency coupling and motivating the two-component focus. Frequency bands for metastability were selected empirically, as principled mappings between oscillatory scales and transformer dynamics remain unclear. Channel selection was random, and although robustness analyses indicate stability, targeted selection strategies may improve sensitivity. Finally, equal weighting of integration and metastability was theoretically motivated but not optimised, and the functional relevance of $\Psi'$ for task performance remains to be established.

Our conditions intentionally induce distinct generative regimes by varying prompt structure, decoding parameters
(temperature/top-$k$), and in some cases model architecture. This design supports discriminability, but it also bundles
factors that may each influence $\Psi'$ (e.g., lexical diversity, output entropy, periodicity of output tokens), limiting
causal attribution to ``reasoning'' per se. Future work should include control conditions that match output entropy
across prompt types, and analyses that account for prompt identity (e.g., prompt-stratified estimates or hierarchical
models with prompt as a random effect).

\subsection{Theoretical and interpretive considerations}

These findings suggest that formal dynamical properties central to neuroscience—hierarchical integration and metastable coordination—also characterise computational regimes in artificial neural systems, supporting the idea of modality-general organisational principles. The component-level dissociations further reinforce theoretical frameworks that treat integration and flexibility as complementary dimensions of organisation \cite{tononi2016integrated,tognoli2014metastable}.

We emphasise that $\Psi'$ does not measure consciousness or subjective experience in artificial systems. It quantifies formal properties of activation dynamics, with higher values indicating more coordinated temporal organisation and lower values reflecting simpler, more rigid dynamics. This distinction is essential for responsible interpretation: while dynamical metrics offer valuable tools for interpretability, they do not warrant anthropomorphic or phenomenological claims about machine experience \cite{dehaene2014consciousness,seth2021being}.

\section{Conclusions}

We adapted established dynamical measures from neuroscience to characterise internal organisation in large language models. The composite index $\Psi'$, combining hierarchical integration and metastability, distinguished structured reasoning from repetitive output, destabilised high-temperature sampling, and two forms of architectural perturbation in GPT-2-medium. These effects were robust to layer-subset recording, random channel subsampling, and multiple random seeds, indicating that $\Psi'$ reflects distributed system-level properties rather than measurement artefacts.

These results constitute a systematic application of composite dynamical metrics to transformer activation time-series and demonstrate that LLM computations exhibit measurable differences in temporal integration and metastable coordination across functional regimes. The observed component-level dissociations further suggest that integration and metastability capture complementary aspects of dynamical organisation.

We emphasise that $\Psi'$ quantifies formal dynamical properties and does not imply subjective experience or consciousness in artificial systems. Instead, elevated $\Psi'$ is interpreted as evidence of more coordinated temporal organisation that may support complex processing.

Several directions remain open. Extending the analysis across model scales, architectures, and training regimes will test generality. Methodologically, longer sequences or alternative estimators may enable incorporation of cross-frequency coupling and fuller comparison to established neuroscience frameworks. Finally, validating $\Psi'$ against behavioural performance, trial-level output quality, targeted ablations, and alternative dynamical measures (e.g., entropy rates, Lyapunov exponents, dimensionality metrics), as well as applying the approach to other sequence domains (e.g., audio, video, multimodal models), will clarify the functional significance of dynamical organisation in artificial neural systems.

\section*{Acknowledgments}

We thank the contributors to the HuggingFace Transformers library and the broader open-source community for developing and maintaining the tools that made this research possible. This work was conducted using open-source software and publicly available pretrained models.

\section*{Author Contributions}
HU and NH conceived the mathematical model. HU developed the algorithms, ran the computational experiments and documented the results. NH provided input to the analysis of the results. Both authors contributed equally to the interpretation of results and manuscript preparation.

\section*{Funding}
No funding was received for this work.

\section*{Availability of data and materials}
Implementation of the base consciousness index $\Psi$, is publicly available on GitHub at: 
\begin{center}
\url{https://github.com/ugail/Index-for-Consciousness-Dynamics}
\end{center}
The code is provided under an open-source license and can be utilised to run the experiments discussed in this paper.

\end{document}